\documentclass[lettersize,journal]{IEEEtran}
\usepackage{amsmath,amsfonts}
\usepackage{array}
\usepackage[caption=false,font=normalsize,labelfont=sf,textfont=sf]{subfig}
\usepackage{textcomp}
\usepackage{stfloats}
\usepackage{url}
\usepackage{verbatim}
\usepackage{graphicx}
\usepackage{cite}
\usepackage{times}
\usepackage{epsfig}
\usepackage{amssymb}
\usepackage{multirow}
\usepackage{makecell}
\usepackage{booktabs}
\usepackage[ruled,vlined]{algorithm2e}

\newcommand{\etal}{\textit{et al}.}
\newcommand{\ie}{\textit{i}.\textit{e}.}
\newcommand{\eg}{\textit{e}.\textit{g}.}

\begin{document}

\title{Revisiting Stereo Triangulation in UAV Distance Estimation}

\author{Jiafan~Zhuang,
        Duan~Yuan,
        Rihong~Yan,
        Weixin~Huang,
        Wenji~Li
        and~Zhun~Fan,~\IEEEmembership{Senior Member,~IEEE,}% <-this % stops a space
\thanks{Jiafan Zhuang, Zhun Fan, Weixin Huang and Wenji Li are with the Department of Electronic Engineering, Shantou University, Shantou, Guangdong 515063, China, and also with the Key Lab of Digital Signal and Image Processing of Guangdong Province, Shantou, Guangdong 515063, China (e-mail: jfzhuang@stu.edu.cn;  zfan@stu.edu.cn;weixinhuang@stu.edu.cn;liwj@stu.edu.cn).
Duan Yuan, Rihong Yan  are with the College of Electronic Engineering, Shantou University, Shantou 515063, Guangdong, China(e-mail: 21dyuan@stu.edu.cn; 21rhyan1@stu.edu.cn)
Corresponding author:~Zhun~Fan\protect\\}
}



\maketitle

\begin{abstract}
Distance estimation plays an important role for path planning and collision avoidance of swarm UAVs. However, the lack of annotated data seriously hinders the related studies. In this work, we build and present a UAVDE dataset for UAV distance estimation, in which distance between two UAVs is obtained by UWB sensors. During experiments, we surprisingly observe that the stereo triangulation cannot stand for UAV scenes. The core reason is the position deviation issue due to long shooting distance and camera vibration, which is common in UAV scenes. To tackle this issue, we propose a novel position correction module, which can directly predict the offset between the observed positions and the actual ones and then perform compensation in stereo triangulation calculation. Besides, to further boost performance on hard samples, we propose a dynamic iterative correction mechanism, which is composed of multiple stacked PCMs and a gating mechanism to adaptively determine whether further correction is required according to the difficulty of data samples. We conduct extensive experiments on UAVDE, and our method can achieve a significant performance improvement over a strong baseline (by reducing the relative difference from 49.4\% to 9.8\%), which demonstrates its effectiveness and superiority. The code and dataset are available at https://github.com/duanyuan13/PCM. 
\end{abstract}


\section{Introduction} \label{sec:introduction}
\IEEEPARstart{R}{ecently}, the research of swarm unmanned aerial vehicles (UAV) has attracted increasing attention from researchers because of its wide applicability, such as environment exploration~\cite{huang2022swarm}, autonomous search and rescue~\cite{kakaletsis2021safety}, target tracking and entrapping~\cite{Li2022Swarm}, flocking~\cite{schilling2021vision}, task allocation~\cite{dong2017formation}, \emph{etc.} To achieve an effective collaboration of swarm UAVs and avoid collisions, a reliable and accurate distance estimation of surrounding UAVs plays an important role.

Stereo-based distance estimation has been widely studied for many decades, especially in autonomous driving~\cite{geiger2013vision, yang2019driving} and in-door robotic applications~\cite{gupta2017indoor, kriegman1989stereo, burri2016euroc}. 
Existing methods~\cite{hirschmuller2007stereo, luo2016efficient,stereo2009, Robust2010Stereo} rely on matching each pixel densely between a pair of images captured by a stereo camera. Typically, a disparity map is predicted by using deep learning techniques and then is transformed into a distance map via stereo triangulation~\cite{mahammed2013object, hartley2003multiple}. Benefited from adequate and high-quality training data~\cite{geiger2012kitti,2018high,2019Tsun-Hsuan,nguyen2022ntu} annotated by using LiDAR, these methods can achieve a promising estimation performance.

\begin{figure}[t]
	\begin{center}
		\includegraphics[width=1.0\linewidth]{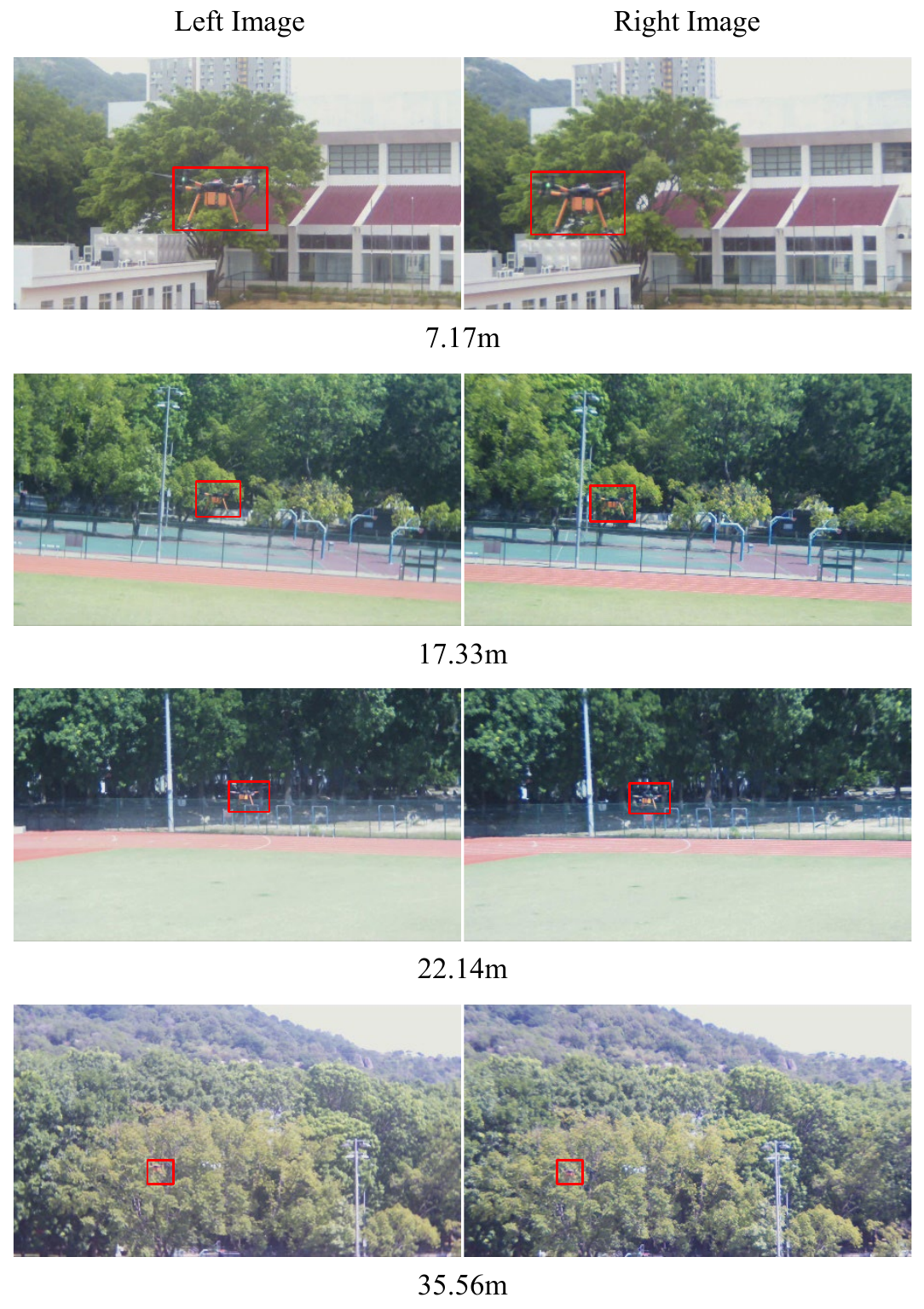}
	\end{center}
	\vspace{-3mm}
	\caption{\textbf{Samples from our presented UAVDE dataset}. We collect thousands of UAV stereo images and annotate them with UAV bounding boxes and distances to the UAV center. Notably, only distance to the UAV center is collected via UWB sensors instead of collecting pixel-wise annotations on LiDAR, which is more economical and efficient for practical UAV applications. Best viewed in color.}
	\vspace{-3mm}
	\label{dataset}
\end{figure}

Distance estimation has been widely used in other scenes, but it is rarely studied in UAV scenes because of two major challenges.
(1) \emph{Data annotation in UAV scenes is difficult.} As a common tool to obtain annotations for distance estimation, LiDAR can only produce sparse point clouds, which is adequate to capture various objects in urban~\cite{niemeyer2014contextual} or in-door scenes~\cite{kumar2017lidar}. However, it is unable to accurately scan a small UAV (\eg, $0.3m$) in a typically long distance (\eg, $30m$)~\cite{hammer2018lidar}.
(2) \emph{Computation resource in UAV system is limited.} Existing estimation methods rely on dense disparity prediction, which is computationally expensive. For example, a popular method AANet~\cite{xu2020aanet} can achieve nearly real-time performance (\eg, $16.13$ FPS) on an NVIDIA V100 device. However, a typical computing device (\eg, NVIDIA Jetson TX2 or Jetson AGX Xavier) on UAVs can only provide around 1/10 or even lower computation capability compared to an NVIDIA V100.

Essentially, these two difficulties are both caused by the commonly adopted estimation paradigm, \ie, dense disparity prediction~\cite{kendall2017end,chang2018pyramid,liang2018learning}. However, \emph{it is not necessary to predict the distance of each pixel in practical UAV scenes.} As shown in Fig.~\ref{dataset}, the interested UAV usually only occupies a small portion in the image. Besides, considering the typically small size, it is adequate for path planning and collision avoidance even only provided with the distance to the UAV center, which is verified in previous successful applications~\cite{mahammed2013object, haseeb2018disnet, bertoni2019monoloco}.

Motivated by this observation, in this work, we build and present a dataset specifically for \textbf{UAV} \textbf{D}istance \textbf{E}stimation, named UAVDE dataset. Different from existing datasets, we only annotate the distance between two UAV centers (\ie, the target UAV and the camera-shooting UAV) rather than densely annotating on each pixel, as shown in Fig.~\ref{dataset} Specifically, we equip ultra-wideband (UWB) sensors on the center of each UAV, which can directly obtain the ground truth distances.

\begin{figure}[t]
	\begin{center}
		\includegraphics[width=1.0\linewidth]{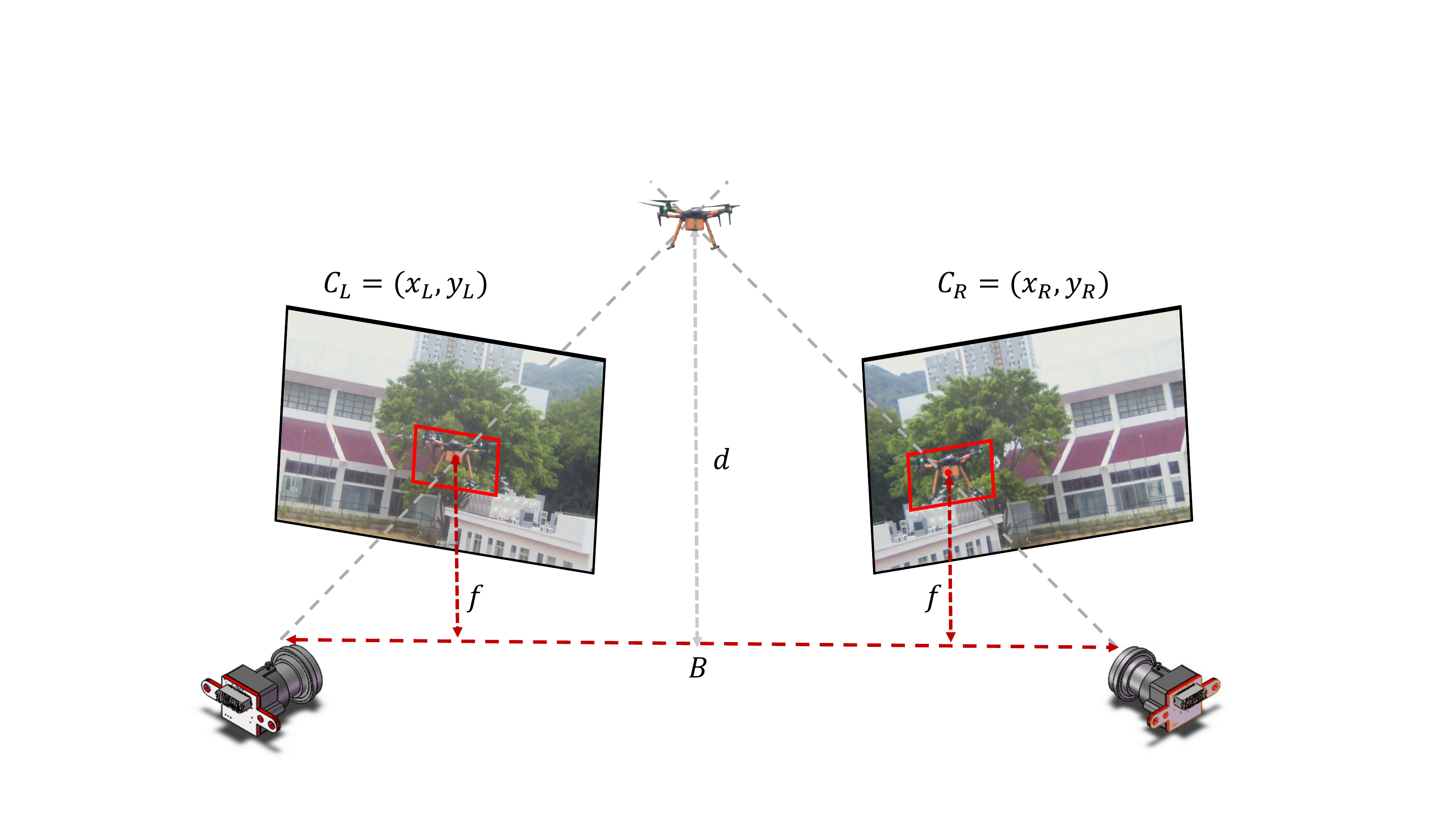}
	\end{center}
	%\vspace{-3mm}
	\caption{\textbf{Stereo triangulation in UAV scenes}. The center of UAVs in stereo images can be obtained by UAV detection, based on which stereo triangulation can be computed to estimate the UAV distance. Here, $B$ and $f$ are the baseline and focal length of the stereo camera, respectively.}
	%\vspace{-3mm}
	\label{pipeline}
\end{figure}

\begin{figure}[t]
	\begin{center}
		\includegraphics[width=1.0\linewidth]{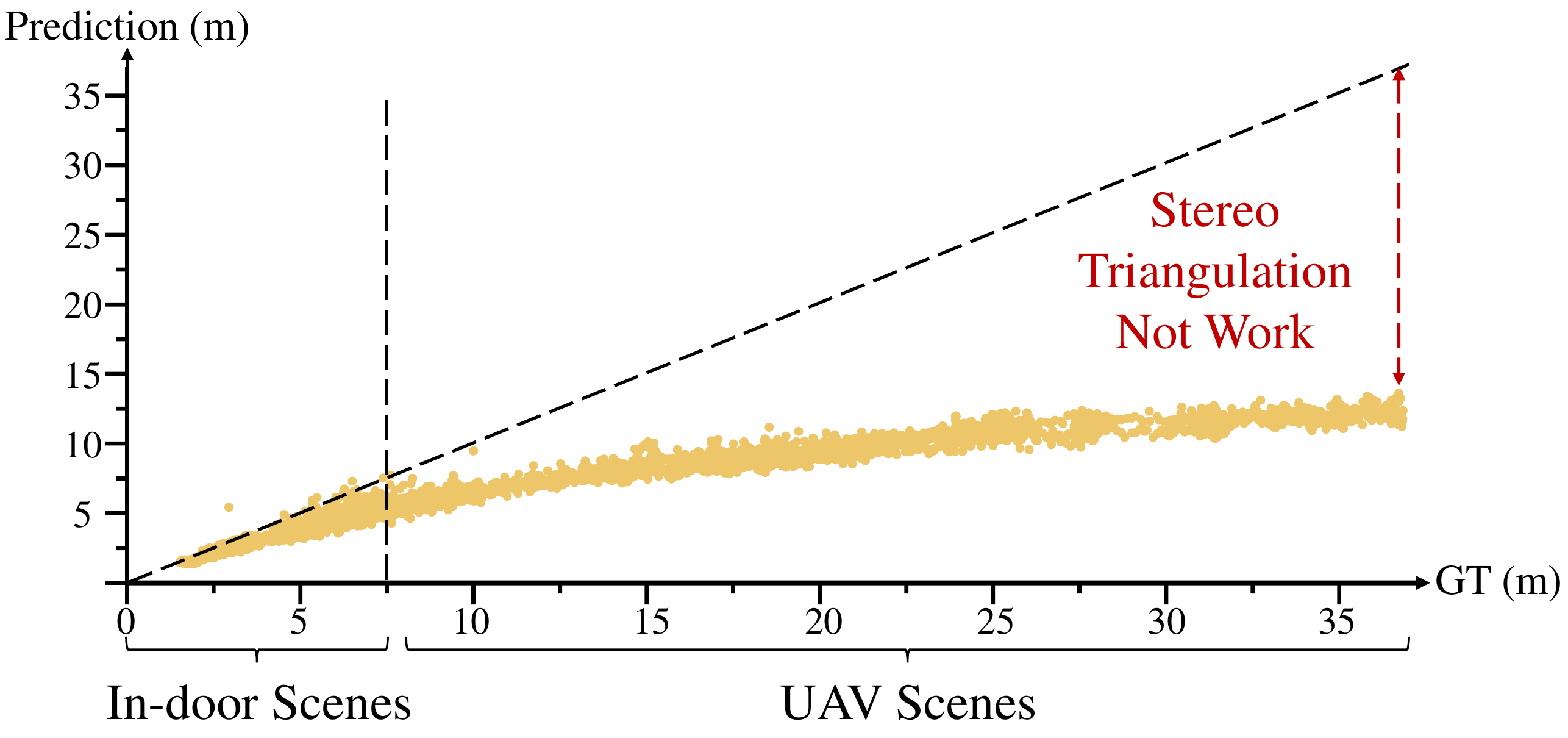}
	\end{center}
	%\vspace{-3mm}
	\caption{\textbf{Upper-bound analysis of stereo triangulation in UAV scenes}. The prediction is relatively accurate when the distance is small (\eg, less than 8 m), but as the distance increases, the prediction gradually deviates from the ground truth.}
	%\vspace{-3mm}
	\label{analysis}
\end{figure}

Based on our proposed dataset, existing distance estimation methods are not applicable due to the lack of dense distance annotations. Following a previous successful work~\cite{mahammed2013object}, a practical solution is to conduct stereo triangulation based on UAV centers from stereo images, as shown in Fig.~\ref{pipeline}. The UAV distance can be estimated as
\begin{equation}
	d=\frac{Bf}{x_{L}-x_{R}},
	\label{st}
\end{equation}
where $B$ and $f$ are the baseline and focal length of the stereo camera, respectively.
The UAV centers can be simply obtained after UAV detection. To validate the effectiveness of this solution, we perform an upper-bound analysis by estimating distance with ground truth bounding boxes, which can remove the interference of the UAV detection performance. However, the experimental results are rather surprising, as shown in Fig.~\ref{analysis}. The prediction is relatively accurate when the distance is small (\eg, $<8m$), but as the evaluating distance increases, the prediction gradually deviates from the ground truth. The results indicate that stereo triangulation does not work in typical UAV scenes (\eg, $>20m$).

Why does stereo triangulation work successfully in indoor applications but not in UAV scenes? Based on previous analysis works~\cite{woods1993distortion, zhang2020depth, gallup2008variable, liang2014robust, szeliski1997shape}, we claim that there are three main reasons:
\begin{itemize}
    \item \emph{A small baseline-to-depth ratio.} The small size of UAVs results in a smaller baseline-to-depth ratio, leading to a narrow triangulation angle for estimating 3D points from image and physical correspondences. This ill-conditioned geometric setting makes accurate estimation challenging~\cite{zhang2020depth, gallup2008variable}.
    \item \emph{A large focal length.} To achieve precise target detection, the focal length is set to an optimal value, which is commonly large in UAV scenes. However, increasing the focal length decreases the field of view angle of the telephoto imaging system, causing the imaging beam to converge closer to the optical axis. This amplifies calibration errors arising from image points, imaging models, and control points, degrading the performance of chessboard-based calibration methods\cite{liang2014robust, zhang2000flexible}.
    \item \emph{Vibrations on camera systems.} UAVs often experience vibrations or shaking during flight due to movement and external factors like wind. These vibrations cause noticeable changes in the images and can turn the stereo frame of the drone from rigidity to non-rigidity\cite{zhang2020depth, szeliski1997shape}.
\end{itemize}

\begin{table}[!t]
	%\vspace{1mm}
	\caption{\textbf{Performance comparison based on stereo images calibrated by different methods}. The performance is still unsatisfactory since the position deviation issue can not be simply tackled by careful camera calibration. Here, Abs Rel and Sq Rel are two common evaluation metrics for distance estimation.}
	%\vspace{-3mm}
	\centering
	\renewcommand{\arraystretch}{1.3}
	\setlength{\tabcolsep}{1mm}
	\begin{tabular}{c|c|c|c|c}
		\toprule
		Methods      & Uncalibrated    & Zhang~\cite{zhang2000flexible}    & Yan \etal~\cite{yan2022opencalib}       & Sch\"{o}ps \etal~\cite{schops2020having}  \\
		\hline\hline
		Abs Rel           & 0.478             & 0.468                              & 0.469                                   &0.471            \\
        Sq Rel        & 5.483             & 5.342                              & 5.349                                   &5.421          \\
        \bottomrule
	\end{tabular}
	\label{calibration}
\end{table}

Then a natural question arises: can this position deviation issue be simply tackled by careful camera calibration? Here, we provide estimation performance on different stereo images, which are calibrated via traditional methods~\cite{zhang2000flexible, yan2022opencalib} and recent learning-based methods~\cite{schops2020having}, respectively. As shown in TABLE~\ref{calibration}, the estimation performance is still unsatisfactory after calibration. This experiment demonstrates that \emph{the position deviation issue is not trivial and needs to be specifically addressed}.

Based on the above analysis, we argue that the deviation of the actual UAV position is the main reason to the estimation error of stereo triangulation. To tackle this issue, we propose a novel method named \textbf{P}osition \textbf{C}orrection \textbf{M}odule (PCM). The main idea is to directly predict the offset between the observed and the actual positions of the target UAV. The predicted offset is then used for calculation compensation in stereo triangulation. Besides, although PCM can effectively alleviate the deviation issue, during experiments we found that some hard samples with serious deviations can not be completely corrected. Therefore, to further boost the performance, we design a \textbf{D}ynamic \textbf{I}terative \textbf{C}orrection (DIC) mechanism. Specifically, we stack multiple PCMs sequentially and design a gating mechanism to adaptively determine whether a further correction is required according to the difficulty of data samples.

We experimentally evaluate the proposed method on the UAVDE dataset. The results validate the effectiveness of our correction method in UAV distance estimation, and it can bring a significant performance improvement. The contribution of this work are summarized as follows.
\begin{itemize}
	\item We formulate the UAV distance estimation task and present a UAVDE dataset.
	\item We discover that the position deviation issue is the key reason to the failure of stereo triangulation in UAV scenes.
	\item We propose a novel position correction module (PCM) and a dynamic iterative correction (DIC) mechanism to accurately predict the offset between the observed and actual positions, which is used for compensation in stereo triangulation calculation.
	\item We experimentally evaluate our proposed method on the UAVDE dataset, which demonstrates the effectiveness and superiority of our method.
\end{itemize}

The rest of this paper is organized as follows. We review the related works on UAV perception and stereo distance estimation in Section~\ref{sec:related}. Section~\ref{sec:dataset} and Section~\ref{sec:method} provide the details of our proposed dataset and approaches, respectively. Section~\ref{sec:exp} experimentally evaluates the proposed method. Finally, we conclude the work in Section~\ref{sec:conclusion}.

\section{Related Work} \label{sec:related}
In this section, we review the literature relevant to our work concerned with visual perception in UAV scenes and common stereo distance estimation methods.

\subsection{UAV Perception}
Drones offer unique perspectives that traditional methods can’t achieve, thereby broadening the application of computer vision in aerial scenarios. This includes the enhancement of datasets~\cite{ding2022object,zhu2022detection} and tasks such as object detection~\cite{deng2021detection}, tracking~\cite{li2023tracking}, saliency prediction~\cite{fu2020saliency}, and scene recognition~\cite{bi2021reco} in UAV scenes. Specifically, Deng~\cite{deng2021detection} introduced an end-to-end global-local self-adaptive network to tackle the issue of unevenly distributed and small-scale objects in drone-view detection. Li~\cite{li2023tracking} achieved superior performance in robust object tracking by focusing on local parts of the object target. Fu~\cite{fu2020saliency} proposed a comprehensive video dataset for aerial saliency prediction. Bi~\cite{bi2021reco} achieved impressive results by extracting key local regions based on a local semantic-enhanced ConvNet.

Although great progress has been made in some visual perception tasks, distance estimation in UAV scenes is still rarely studied, despite the fact that it is crucial for UAV applications. In this work, we formulate the UAV distance estimation task and present a well-established dataset to aid the corresponding researches.

\subsection{Classical Stereo Matching}
For depth estimation from stereo images, many methods have been proposed in the literature, which consist of matching cost computation and cost volume optimization. According to ~\cite{scharstein2002middlebury}, a classical stereo matching algorithm consists of four steps: matching cost computation, cost aggregation, optimization, and disparity refinement. As the pixel representation plays a critical role in the process, previous literature has exploited a variety of representations, from the simplest RGB colors to hand-craft feature descriptors~\cite{yoo2009fast,mouats2015repre,mozerov2015stereo,lowe1999object,bay2006surf}.
Together with postprocessing techniques like Markov random fields~\cite{szeliski2006comparative}, semi-global matching~\cite{hirschmuller2007stereo} and Bayesian approach~\cite{geiger2010efficient}, these methods can work well on relatively simple scenarios, such as in-door scenes. 

However, although most classical methods do not require a density map annotation, these methods often encounter challenges in real-world scenarios such as occlusions~\cite{yan2019occlusion, yang2008stereo}, diverse lighting conditions~\cite{hirschmuller2009lighting}, and featureless regions~\cite{hirschmuller2007stereo}, which are commonly observed in UAV scenes.

\subsection{Learning-based Stereo Matching}
To deal with more complex real-world scenes, recent researchers leverage deep-learning techniques to extract pixel-wise features and match correspondences~\cite{ji2017surfacenet,zbontar2016stereo,kendall2017end,chang2018pyramid,liang2018learning,khamis2018stereonet,hartmann2017learned}. The learned representation shows more robustness to low-texture regions and various lightings~\cite{huang2018deepmvs,yao2018mvsnet,yao2019recurrent,chen2019point,luo2019p}. Rather than directly estimating depth from image pairs, some approaches also tried to incorporate semantic cues and context information in the cost aggregation process~\cite{yang2018segstereo,cherabier2018learning,im2019dpsnet,zeng2022cost} and achieved positive results. 

Although learning-based methods can achieve significant improvement on estimation accuracy, they all rely on adequate and high-quality training data~\cite{geiger2013vision,cheng2022branch,zeng2022cost,geyer2020a2d2} densely annotated by LiDAR. However, LiDAR can not be applied in UAV scenes~\cite{farlik2016radar, hammer2018lidar}, and thus can not provide critical dense annotations for existing learning-based methods. Different from existing works, we build a new dataset for UAV distance estimation, which obtains UAV distance based on UWB sensors instead of LiDAR. Based on this dataset, we revisit the stereo triangulation solution and discover the key issue of UAV distance estimation, \ie, position deviation, and propose a novel position correction method.

\section{UAVDE Dataset} \label{sec:dataset}
To aid the study of stereo distance estimation in UAV scenes, we present a novel \textbf{UAV} \textbf{D}istance \textbf{E}stimation (UAVDE) dataset. Here, we would introduce the data collection process and the details of the annotation process.

\subsection{Data Collection}
\begin{figure}[t]
	\begin{center}
		\includegraphics[width=0.75\linewidth]{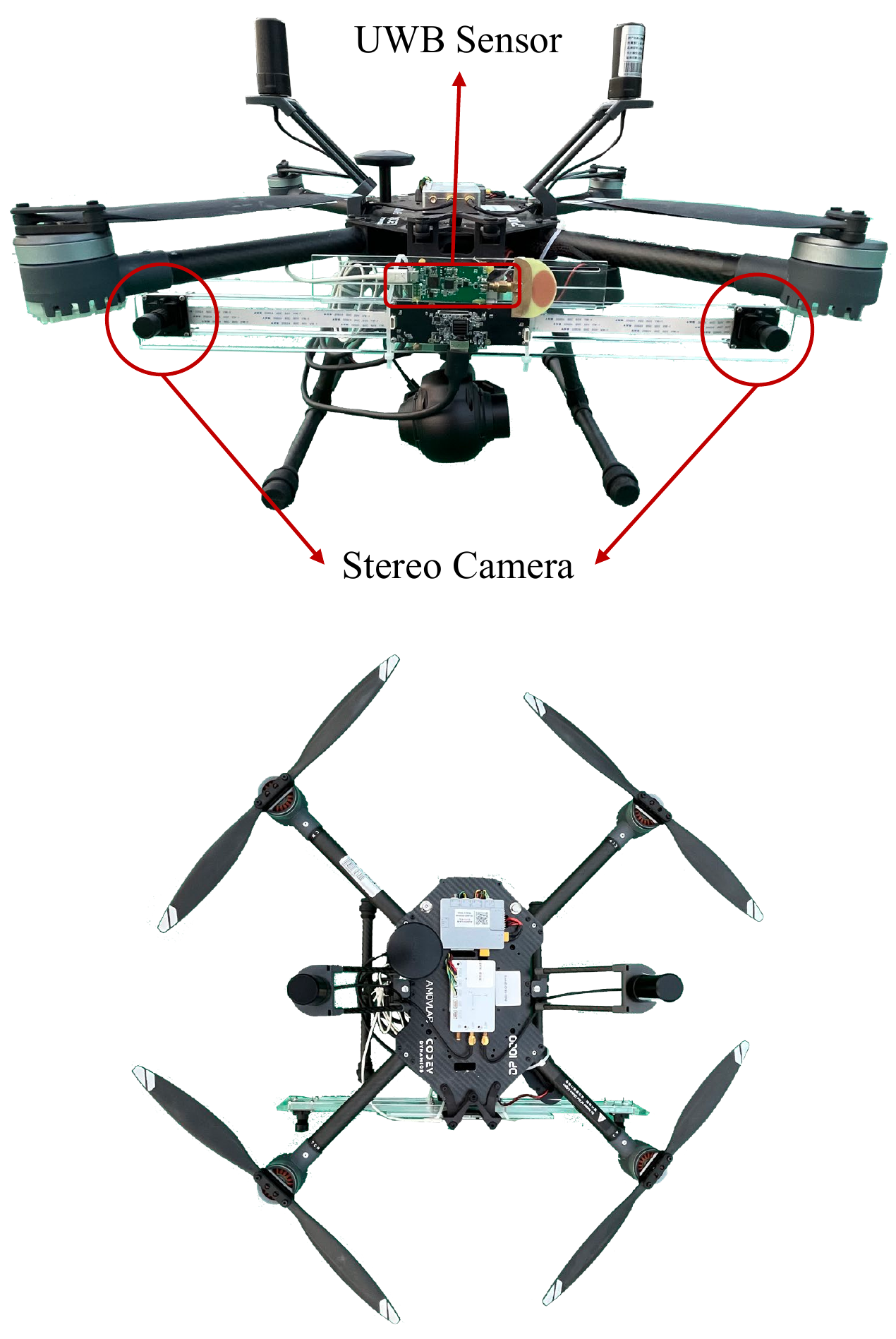}
	\end{center}
	%\vspace{-3mm}
	\caption{\textbf{The illustration of the recording UAV}. The UAV is equipped with a stereo camera for image data collection and a UWB sensor for distance annotation collection.}
	%\vspace{-3mm}
	\label{uav}
\end{figure}

To collect stereo images, we particularly use a recording UAV and a target UAV. The recording UAV is an AMOV P600 with a mounted stereo camera and an ultra-wideband (UWB) sensor, as shown in Fig.~\ref{uav}. Limited by the size of the UAV, the baseline of the stereo camera is set to $406mm$. The equipped lens have a focal length of $12mm$, while the horizontal and vertical FOV are $22^\circ$ and $18^\circ$, respectively.
Besides, it also equips an NVIDIA Jetson Xavier NX as computing platform, which is used for practical evaluation of distance estimation techniques. The target UAV is a DJI M200 with a compact shape, which can achieve fast and steady flight performance and is suitable to serve as a detected UAV.

\begin{figure}[!t]
	\begin{center}
		\includegraphics[width=1.0\linewidth]{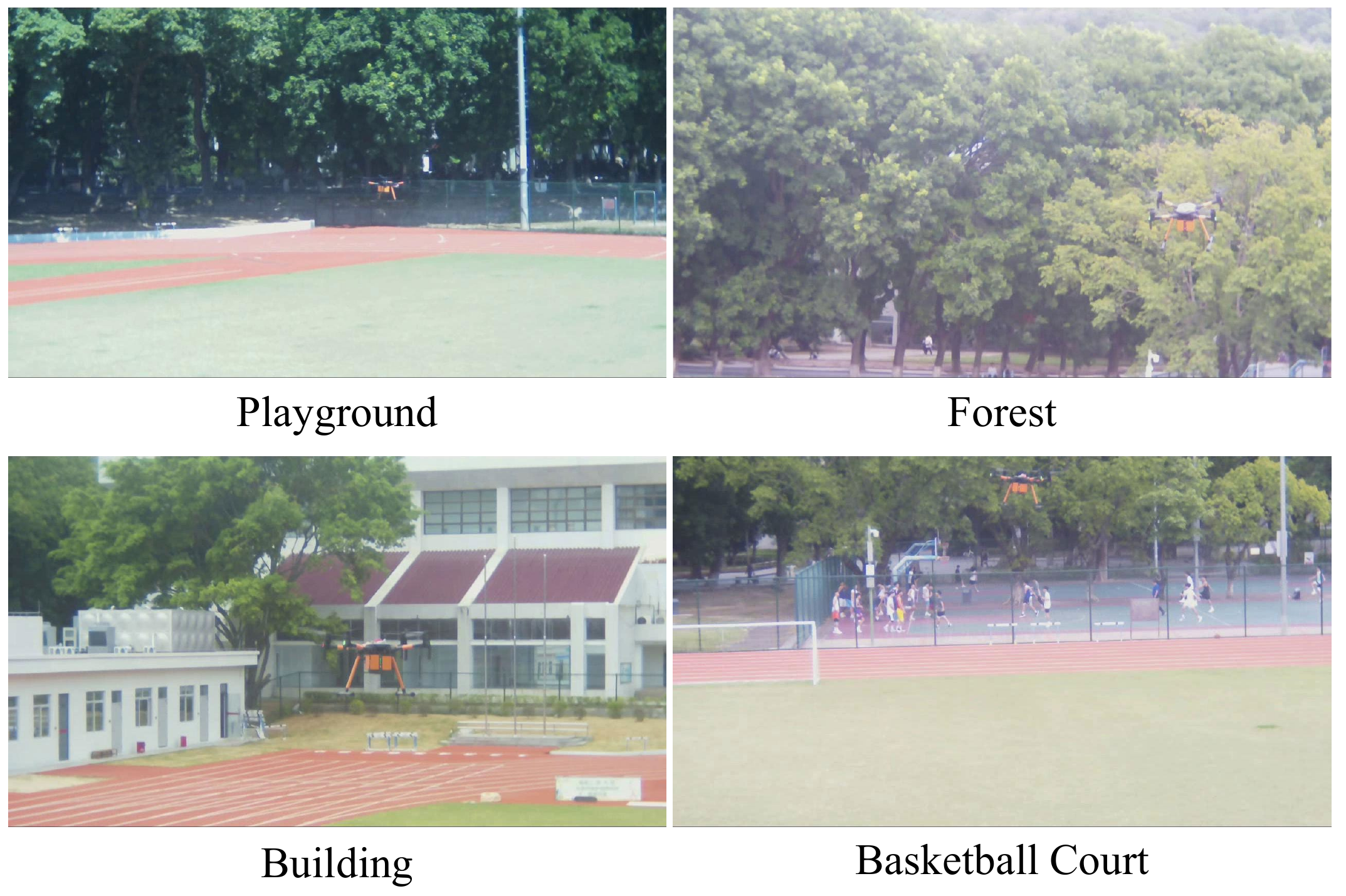}
	\end{center}
	%\vspace{-3mm}
	\caption{\textbf{The illustration of typical UAV scenarios}. We collect data in typical UAV scenarios, \eg, playground, forest, building and basketball court.}
	%\vspace{-3mm}
	\label{scene}
\end{figure}

Motivated by practical applications, we select several typical scenarios for data collection, \eg, buildings, forests, playgrounds, and basketball courts, as shown in Fig.~\ref{scene}. Based on these scenes, we have totally collected 3895 stereo images, which are divided into the training, validation, and evaluation subsets. 
To evaluate the adaptation ability to unseen scenes, the training subset contains different scenarios from the others when splitting the dataset.

\subsection{Data Annotation}
Different from existing datasets with dense distance annotations by LiDAR, we propose to collect the distance on the center of the target UAV by considering the requirements of practical applications. Here, we particularly use UWB sensors for distance annotation, since UWB positioning technique can precisely measure distance by calculating the time it takes for signals to travel amongst the sensors on UAVs. Comparing to LiDAR, our annotation pipeline is rather economical and efficient, which can be easily extended to new scenarios and UAVs.
Besides, we manually annotate the UAV bounding boxes on stereo images, which is necessary for UAV detector training.

\section{Method} \label{sec:method}

\begin{figure}[!t]
	\begin{center}
		\includegraphics[width=1.0\linewidth]{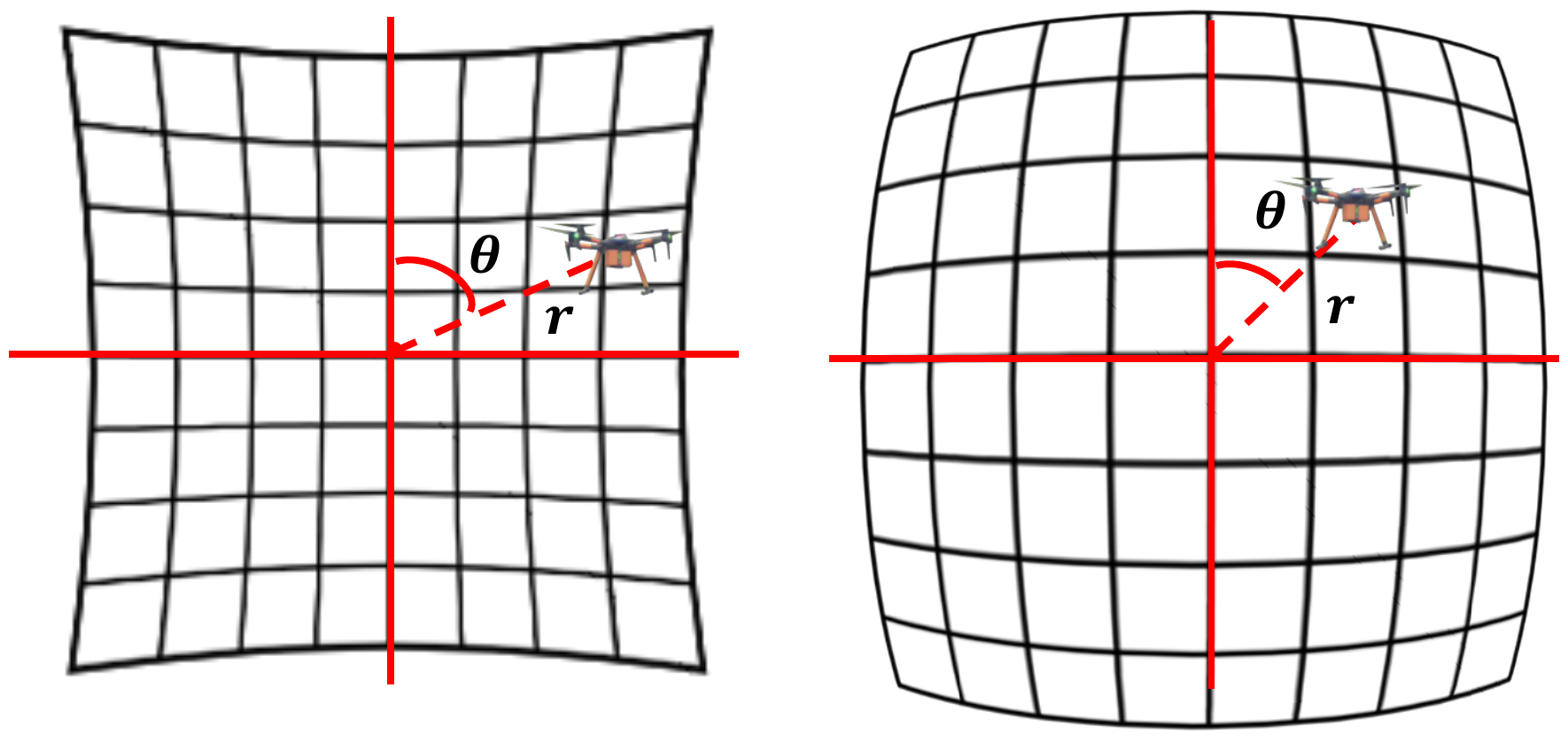}
	\end{center}
	%\vspace{-3mm}
	\caption{\textbf{Illustration of typical pincushion and barrel distortion phenomenons}. Intuitively, the severity of position deviation is related to the image position of UAVs. Best viewed in color.}
	%\vspace{-3mm}
	\label{distortion}
\end{figure}

\subsection{Position Correction Module}
In this work, we mainly focus on addressing the position deviation issue and propose a novel position correction module (PCM) to explicitly predict the offset between the actual and deviated position of the target UAV. To this end, we first need to figure out what factors are highly related to the position deviation issue. Fig.~\ref{distortion} illustrates two typical image distortion phenomenons, \ie, pincushion and barrel distortion. It can be intuitively observed that the severity of position deviation is related to the image position, which can be represented by the relative angle $\theta$ and radius $r$ to the image center. Similar conclusion is also illustrated in previous works~\cite{zhang2013object, ding2019learning}. Besides, as shown in Fig.~\ref{analysis}, position deviation is also affected by the distance of target UAV. Since the distance is unavailable, we can simply use the size (\ie, $w$ and $h$) of detected UAV bounding boxes as a rough representation.

\begin{figure}[!t]
	\begin{center}
		\includegraphics[width=0.8\linewidth]{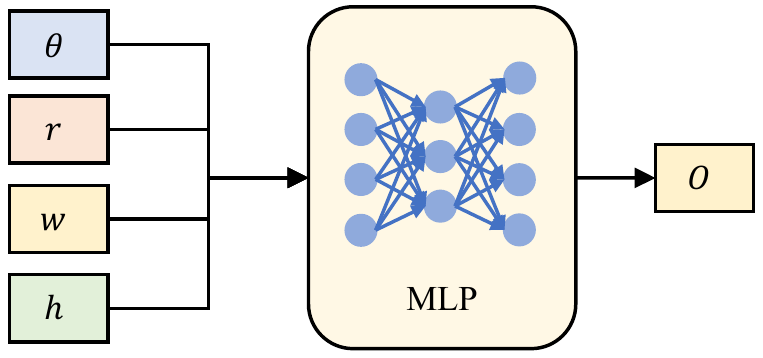}
	\end{center}
	%\vspace{-3mm}
	\caption{\textbf{Illustration of Position Correction Module (PCM)}. Given several parameters of the target UAV position, we use a simple MLP to predict the offset between the actual and deviated position. Best viewed in color.}
	%\vspace{-3mm}
	\label{pcm}
\end{figure}

Based on the above analysis, we propose to use a 4-tuple $\{\theta, r, w, h\}$ to predict the position offset, as shown in Fig.~\ref{pcm}. In the position correction module, a simple multilayer perceptron (MLP) is adopted to perform prediction. Theoretically, any off-the-shelf MLP model can be adopted. Since the stereo triangulation is only related to the horizontal coordinate in Eq.~\ref{st}, the MLP is designed to regress a single offset variable $O$ along the X-axis. Practically, PCM is conducted on left and right images for correction respectively, which results in $O_{L}$ and $O_{R}$. Based on the predicted offsets, we can 
compensate the computation of stereo triangulation as

\begin{equation}
	\tilde{d}=\frac{Bf}{(x_{L}+O_{L})-(x_{R}+O_{R})}.
\end{equation}

During training, a simple L2 loss is applied on the corrected distance $\tilde{d}$ with the ground truth distance $d_{gt}$. 
\begin{equation}
	L_{PCM}=(\tilde{d}-d_{gt})^{2}.
\end{equation}
Notably, the training of UAV detector and PCM is completely disentangled. Any off-the-shelf detector can be pre-trained on the UAVDE dataset and then used to produce 4-tuples $\{\theta, r, w, h\}$ for PCM training. During inference, PCM is simply attached to the end of UAV detector for position correction.

%-------------------------------------------------------------------------
\subsection{Dynamic Iterative Correction}

\begin{figure*}[t]
	\begin{center}
		\includegraphics[width=1.0\linewidth]{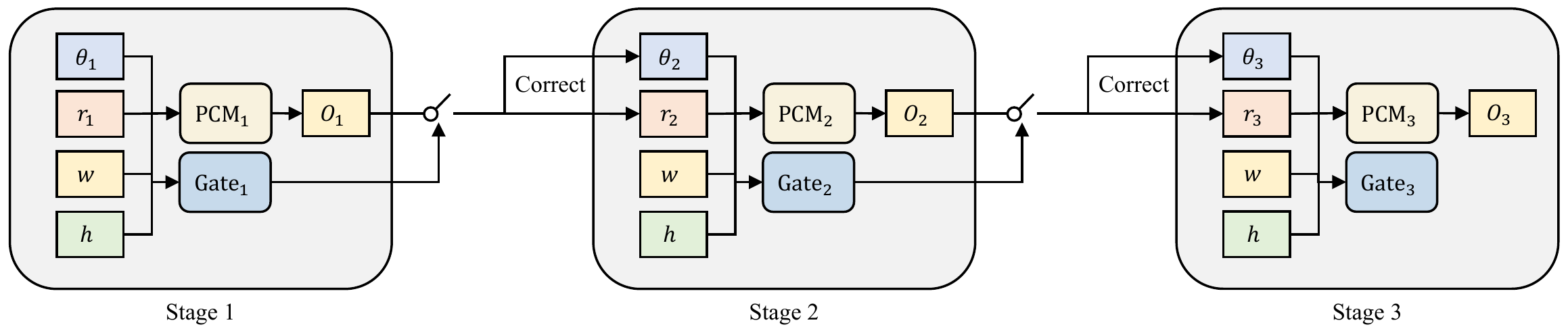}
	\end{center}
	%\vspace{-3mm}
	\caption{\textbf{Illustration of dynamic iterative correction (DIC)}. To tackle hard samples, we propose to perform iterative correction stages by stacking multiple PCMs sequentially, which is adaptively adjusted by our proposed gating module according to the difficulty of samples.}
	%\vspace{-3mm}
	\label{dynamic}
\end{figure*}

Although PCM can achieve effective correction and significant performance improvement, during experiments we found that some hard samples with large position deviation can not be completely corrected. To further boost the performance, we propose to perform iterative correction on hard samples by stacking multiple PCMs sequentially. Here, the key issue is to determine whether the input sample is required to conduct further correction. Motivated by successful practices~\cite{li2021dynamic,han2021dynamic} in dynamic network, we design a simple but effective gating mechanism to adaptively adjust the correction procedure according to the difficulty of data samples.

Specifically, we implement the gate module with another independent MLP, which also takes the 4-tuple as input and outputs an indicator as a switch to determine whether to perform further correction, as shown in Fig.~\ref{dynamic}. During training, we first determine whether the input sample belonging to hard samples according to its distance estimation error rate after the first PCM. Here, we adopt a commonly used evaluation metric, \ie, absolute relative difference ($Abs Rel$), for measurement.
\begin{equation}
	err=eval(\tilde{d}, d_{gt}),
\end{equation}
If the error rate is larger than a pre-defined threshold $T$, we can regard the sample as a hard sample. 
\begin{equation}
	\Phi(\mathcal{X})=\mathbb{I}(err>T)
	\label{threshold}
\end{equation}
Here, $\mathcal{X}$ is the input 4-tuple, $\mathbb{I}$ is an indicator function and $\Phi(\mathcal{X})\in\{0,1\}$ represents whether $\mathcal{X}$ is a hard sample. After that, we impose a cross-entropy loss on the output of the gate module:
\begin{equation}
	L_{Gate}=L_{CE}(Gate(\mathcal{X}), \Phi(\mathcal{X})),
\end{equation}
where $L_{CE}(x,y)=-\sum y*\log (x)$ is the Cross-Entropy loss over softmax activated gate scores and the generated gate target. Based on this implementation, it is easy to extend to more correction stages with multiple PCMs and gate modules:
\begin{equation}
	L_{Gate_{i}}=L_{CE}(Gate_{i}(\tilde{\mathcal{X}}_{i}), \Phi(\tilde{\mathcal{X}}_{i})),
\end{equation}
\begin{equation}
	\tilde{d_{i}}=\frac{Bf}{(x_{Li}+O_{Li})-(x_{Ri}+O_{Ri})}.
\end{equation}
\begin{equation}
	L_{PCM_{i}}=(\tilde{d_{i}}-d_{gt})^{2},
\end{equation}
where$\tilde{\mathcal{X}}_{i}$ represents the $i$th corrected 4-tuple.
Finally, the overall training objective can be presented as
\begin{equation}
	L=\sum_{i=1}^{N-1} (L_{PCM_{i}}+\lambda L_{Gate_{i}}),
	\label{loss}
\end{equation}
where $N$ is the number of correction stages and $\lambda$ is a trade-off parameter. During inference, multiple PCMs are executed sequentially based on instructions from the corresponding gating modules.

\begin{algorithm}[t]
  \caption{UAV Distance Estimation}
  \label{alg:pipeline}
  \KwIn{A Stereo Image: $I_{l}$ and $I_{r}$}  
  \BlankLine
  \textbf{UAV Detection} \\
  \For{$i\in\{l, r\}$}{
      $\{x_{i}, y_{i}, w_{i}, h_{i}\}$ = Detector($I_{i}$) \\
      $\{\theta_{i}, r_{i}\}$ = $RC(x_{i}, y_{i})$ \\
      $\mathcal{X}_{i}$ = $\{\theta_{i}, r_{i}, w_{i}, h_{i}\}$ \\
      }
      \BlankLine
  \textbf{Position Offset Prediction} \\
  $stage$ = 0 \\
  \For{$i\in\{l, r\}$}{
      $\tilde{x}_{i} = x_{i}$ \\
    }
  \While{$stage < N$}{
     \For{$i\in\{l, r\}$}{
        $O_{i}$ = $PCM_{stage}(\mathcal{X}_{i})$ \\
        $\tilde{x}_{i} = \tilde{x}_{i} + O_{i}$ \\
        }
     $t = (Gate_{stage}(\mathcal{X}_{l}) + Gate_{stage}(\mathcal{X}_{r}))/2$  \\
     \If{$t < T$}{
        break \\
     }
    \For{$i\in\{l, r\}$}{
      $\{\tilde{\theta}_{i}, \tilde{r}_{i}\}$ = $RC(\tilde{x}_{i}, y_{i})$ \\
      $\mathcal{X}_{i}$ = $\{\tilde{\theta}_{i}, \tilde{r}_{i}, w_{i}, h_{i}\}$ \\
    }
    $stage = stage + 1$ \\
    }
  \BlankLine
  $d$ = Stereo Triangulation($\tilde{x}_{l}$, $\tilde{x}_{r}$) \\
  \KwOut{A Estimated Distance: $d$}
\end{algorithm}

%-------------------------------------------------------------------------
\subsection{Entire Pipeline}
After introducing our proposed approaches, here, we provide an overview of the entire pipeline for distance estimation, which is shown in Algorithm~\ref{alg:pipeline}. Specifically, RC represents the radian conversion, which can perform the conversion from the cartesian coordinates into the polar coordinates.

\section{Experiments} \label{sec:exp}

%-------------------------------------------------------------------------
\subsection{Dataset}
In this work, we build and present UAVDE dataset to aid the study on distance estimation in UAV scenes. The training, validation and evaluation subsets contain 2815, 541 and 539 stereo images at a resolution of 1280 $\times$ 720, respectively. Each sample contains a distance annotation and UAV bounding boxes annotations on both left and right images. The validation subset is used for hyper-parameter and model selection. Following previous works~\cite{eigen2014depth, garg2016unsupervised, zhou2021r}, we adopt two commonly used evaluation metrics, \ie,  $Abs Rel$ and $Sq Rel$:
\begin{equation}
	Abs Rel=\frac{1}{N}\sum_{i=1}^{N}\frac{|d^{i}-d^{i}_{gt}|}{d^{i}_{gt}},
\end{equation}
\begin{equation}
	Sq Rel=\frac{1}{N}\sum_{i=1}^{N}\frac{||d^{i}-d^{i}_{gt}||^{2}}{d^{i}_{gt}},
\end{equation}

%-------------------------------------------------------------------------
\subsection{Implementation Details}
In this work, we particularly adopt YOLOX-Nano~\cite{yolox2021} as the UAV detector due to its excellent trade-off between performance and computational efficiency, which is critical for computing device on UAVs. Specifically, we first pretrain the detector on the UAVDE dataset and follow the original training protocol, which achieve $55.3$ mAP on validation subset with a small inference resolution (352, 192) for efficiency. After pretraining, we fix the detector to generate 4-tuples $\{\theta, r, w, h\}$ for training our proposed modules.

For our PCM and the gate module, we implement them with an identical MLP architecture, \ie, MLP-Mixer~\cite{tolstikhin2021mlp}, which is a popular and effective MLP variant among various vision tasks. Since MLP-Mixer is designed for a sequence of embeddings projected from image patches, we can naturally feed our generated 4-tuple $\{\theta, r, w, h\}$ into MLP-Mixer sequentially. Benefited from the mixing mechanism, MLP-Mixer can capture the internal relationship within the 4-tuple and predict the position offset. To further improve efficiency, we cut the original 8-layer MLP-Mixer variant into 2-layer, which is sufficient for offset prediction task and can alleviate the overfitting issue. When training the PCM and gate module, we adopt a SGD optimizer with a learning rate of $10^{-3}$, a momentum of 0.9 and a weight decay of $10^{-3}$. Besides, by following the original training protocol, gradient clipping and a cosine learning rate schedule with a linear warmup are adopted. 

There are two hyperparameters in Eq.~\ref{threshold} and Eq.~\ref{loss}, \ie, $T$ and $\lambda$. Actually, different thresholds for each gate module can be adjusted for better correction effect on hard samples. However, in this work, we meant to introduce the novel position correction mechanism and thus not intend to tune the hyperparameters excessively. Therefore, we simply use an identical threshold $T$ for all gate modules, which can bring significant improvement on hard samples. By validating on the val subset, we practically set $T=0.06$ and $\lambda=1.0$ for all experiments.

%-------------------------------------------------------------------------
\subsection{Performance Comparison}
Since dense distance annotations from LiDAR is unavailable in UAV scenes, existing learning-based methods are not applicable. To demonstrate the superiority of our method, we make a comparison with two popular classical methods, \ie, ADCensus~\cite{mei2011accurate} and ELAS~\cite{geiger2010efficient}, which do not rely on dense annotations. Particularly, we reproduce these methods according to their official codes on the proposed UAVDE dataset for fair comparison. 

\begin{table}[!t]
	%\vspace{1mm}
	\caption{\textbf{Performance Comparison on the UAVDE dataset}. Baseline and Baseline* represent stereo triangulation on UAV detection results and annotated bounding boxes, respectively. Our method significantly outperforms other methods}
	%\vspace{-1mm}
	\centering
	\renewcommand{\arraystretch}{1.3}
	\begin{tabular}{l|c|c|c|c}
		\toprule
		\multicolumn{1}{c|}{\multirow{2}{*}{Method}}	    & \multicolumn{2}{c|}{Val} 				& \multicolumn{2}{c}{Test}					\\
		\cline{2-5}
				          		& Abs Rel			& Sq Rel			& Abs Rel			& Sq Rel	\\
		\hline\hline
		ADCensus~\cite{mei2011accurate} 		& 0.601				& 13.588			& 0.714				& 22.911	        \\
		ELAS~\cite{geiger2010efficient}			& 0.508				& 6.609				& 0.527				& 6.467		        \\
		\hline
		Baseline				                & 0.490				& 6.716				& 0.494				& 6.818		          \\
		Baseline*				                & 0.483				& 6.550				& 0.488				& 6.712		           \\
		Ours					                & \textbf{0.114}	& \textbf{0.673}	& \textbf{0.098}	& \textbf{0.401}		\\
		\bottomrule
	\end{tabular}
	\label{uavde}
\end{table}

The performance comparison is shown in TABLE~\ref{uavde}. Here, Baseline and Baseline* represent stereo triangulation on UAV detection results and annotated bounding boxes, respectively. From the results, we have the following observations. 
First, classical methods perform poorly on UAV scenes. They are affected by environmental interference, which results in error-prone disparity estimation.
Second, baseline performs better than classical methods benefited from the robustness of deep learning model, \ie, the YOLOX. However, it still suffers from the position deviation issue. Annotated bounding boxes can only slightly improve the performance, which indicates that we can not tackle this issue by using a stronger detector.
Third, our proposed method can significantly outperform its counterparts by over $39.0\%$ improvement by compensating the position deviation, which demonstrates its superiority and effectiveness.

%-------------------------------------------------------------------------
\subsection{Ablation Study}
In this subsection, we conduct experiments to reveal the effectiveness of our proposed method.

\subsubsection{Effect of Components}
\begin{table}[!t]
		%\vspace{1mm}
	\caption{\textbf{Ablation study on our proposed components.} Each component can bring performance improvement compared to the baseline.}
	%\vspace{-1mm}
        \centering
	\renewcommand{\arraystretch}{1.3}
	\begin{tabular}{l|c|c|c|c}
		\toprule
		\multicolumn{1}{c|}{\multirow{2}{*}{Method}}  & \multicolumn{2}{c|}{Val} 				& \multicolumn{2}{c}{Test}				 \\
		\cline{2-5}
				          	& Abs Rel			& Sq Rel			& Abs Rel				& Sq Rel 		  \\
		\hline\hline
		Baseline	 			& 0.490				& 6.716				& 0.494					& 6.818			   \\
		+ PCM					& 0.148				& 1.014				& 0.121					& 0.620			   \\
		+ PCM + DIC				& \textbf{0.114}	& \textbf{0.673}	& \textbf{0.098}		& \textbf{0.401}	\\
        + PCM + DIC*			& 0.129				& 0.823				& 0.121					& 0.576	             \\
		\bottomrule
	\end{tabular}

	\label{components}
\end{table}

Here, we conduct an ablation study to reveal the contribution of our proposed components, and the results are shown in TABLE~\ref{components}. It can be seen that PCM can significantly improve distance estimation by tackling the position deviation issue. Besides, DIC can further boost the performance by conducting multiple corrections on hard samples. 

To clearly reveal the effect of DIC, we conduct a comparison experiment by using multiple correction on each sample regardless of its difficulty, \ie, DIC*, which performs worse than DIC. The result demonstrates that performing multiple correction on easy samples leads to the over-correction issue. Therefore, a different treatment based on sample difficulty by DIC is necessary.

\subsubsection{Effect of Number of Correction Stages}
\begin{table}[!t]
	%\vspace{3mm}
	\caption{\textbf{Ablation study on the number of corrections.} '0' represents the baseline without position correction. Performing two corrections can bring significant improvement, while the effect of further corrections is not obvious.}
	%\vspace{-1mm}
	\centering
	\renewcommand{\arraystretch}{1.3}
	\begin{tabular}{c|c|c|c|c}
		\toprule
		\multirow{2}{*}{Number}	& \multicolumn{2}{c|}{Val} 				& \multicolumn{2}{c}{Test}					\\
		\cline{2-5}
				        & Abs Rel			& Sq Rel			& Abs Rel				& Sq Rel 			\\
		\hline\hline
		0	 			& 0.490				& 6.716				& 0.494					& 6.818				\\
		1				& 0.148				& 1.014				& 0.121					& 0.620				\\
		2				& 0.114	            & 0.673	            & 0.098                 & 0.401	            \\
		3				& 0.113	            & \textbf{0.663}	& 0.097					& 0.356				\\
		4				& \textbf{0.112}	& 0.693				& \textbf{0.097}		& \textbf{0.351}				\\
		\bottomrule
	\end{tabular}
	\label{number}
\end{table}

Since DIC can bring further improvement on hard samples, here we conduct an ablation study to reveal the effect of the number of corrections, as shown in TABLE~\ref{number}. Here, '0' represents the baseline without position correction. It can be seen that performing two corrections can bring a significant improvement, while the effect of further corrections is not obvious. Considering the trade-off between performance and resource consumption on the UAV platform, we propose to conduct two correction stages.

\subsubsection{Effect of Threshold $T$}
\begin{table}[!t]
	%\vspace{3mm}
	\caption{\textbf{Ablation study on the threshold $T$ in the gate module.} The study is conducted with two PCMs. Here, setting $T=1.0$ represents not performing the second correction, which is used for comparison. It can be seen that our method is relatively robust to the selection of $T$.}
	%\vspace{-1mm}
	\centering
	\renewcommand{\arraystretch}{1.2}
	\begin{tabular}{c|c|c|c|c}
		\toprule
		\multirow{2}{*}{Threshold}	& \multicolumn{2}{c|}{Val} 				& \multicolumn{2}{c}{Test}		\\
		\cline{2-5}
				          & Abs Rel			& Sq Rel			& Abs Rel				& Sq Rel 			\\
		\hline\hline
		1.0				    & 0.148				& 1.014				& 0.121					& 0.620				\\
		\hline
		0.05	 			& 0.118				& 0.693				& 0.100					& \textbf{0.369}	\\
		0.06				& \textbf{0.114}    & \textbf{0.673}    & \textbf{0.098}		& 0.401         	\\
		0.07				& 0.125	            & 0.775	            & 0.103					& 0.428				\\
		0.08				& 0.116				& 0.697				& 0.102					& 0.428				\\
		0.09				& 0.116				& 0.702				& 0.103					& 0.430				\\
  	0.10				& 0.117				& 0.719				& 0.101					& 0.438				\\
		\bottomrule
	\end{tabular}
	\label{table:threshold}
\end{table}

The threshold $T$ in the gate module can affect the selection of hard samples. Here, we conduct an ablation study to reveal the effect of threshold $T$, as shown in TABLE~\ref{table:threshold}. 
Here, setting $T=1.0$ represents not performing the second correction, which is used for comparison. It can be seen that our method can consistently bring improvements with a large range of $T$, demonstrating its robustness to the selection of $T$. By validating on the val subset, we practically set $T=0.06$ as default.

\subsubsection{Analysis on Resource Consumption}
\begin{table}[!t]
	%\vspace{3mm}
	\caption{\textbf{Ablation study on Resource Consumption.} We analyze the parameters and computation cost of our method. The actual speed are evaluated on an NVIDIA Jetson Xavier NX device. The computation cost of the whole correction pipeline is an average cost based on dynamic behaviors on different samples.}
	%\vspace{-1mm}
	\centering
	\renewcommand{\arraystretch}{1.3}
	\begin{tabular}{c|c|c|c}
		\toprule
		Module	          		& Param (M)			& FLOPs (G)	         & FPS			\\
		\hline\hline
		Detector	 			& 0.90				& 1.08				 & 265			\\
		PCM						& 0.02				& 3.144*10$^{-6}$	 & 1555			\\
		Gate					& 0.02				& 3.144*10$^{-6}$	 & 1555			    \\
		\hline
		Detector+PCM	        & 0.92				& 1.08               & 250			\\
        Detector+PCM+DIC       & 0.96              & 1.08               & 236           \\
		\bottomrule
	\end{tabular}
	\label{computation}
\end{table}

Since the computing resource is highly limited on UAVs, the computation efficiency is critical for UAV distance estimation techniques. Here, we conduct an ablation study resource consumption of our proposed method, as shown in TABLE~\ref{computation}. Obviously, comparing to the detector, the parameter and computation cost of our proposed PCM and gate module is negligible benefited to the extremely light-weight design. Since the correction behavior is dynamic according to the sample difficulty, we calculate the average computation cost of the whole correction pipeline as follow:
\begin{equation}
	C_{mean}=\frac{(C_{PCM}+C_{Gate}) * N_{easy} + C_{PCM}*N_{hard}}{N_{easy} + N_{hard}},
\end{equation}
where $C_{PCM}$ and $C_{Gate}$ represents the computation cost of PCM and the gate module, respectively. $N_{easy}$ and $N_{hard}$ represents the number of easy and hard samples, respectively.

\subsubsection{Effect of DIC on Hard Samples}
\begin{table}[!t]
	%\vspace{3mm}
	\caption{\textbf{Analysis of the effect on hard samples.} We analysis the effect of our PCM and DIC on samples with different difficulty on the test subset of UAVDE. PCM can improve the performance consistently on various samples, while DIC mainly focuses on the further improvement on hard samples, eg, samples with distance larger than 20m.}
	%\vspace{-1mm}
	\centering
	\renewcommand{\arraystretch}{1.3}
	\begin{tabular}{l|c|c|c|c}
		\toprule
		\multicolumn{1}{c|}{\multirow{2}{*}{Method}} & \multicolumn{2}{c|}{Val}    & \multicolumn{2}{c}{Test}				             \\
		\cline{2-5}
				 		  & $<20m$			  & $>20m$		    & $<20m$		   & $>20m$	        	\\
		\hline\hline
		Baseline	 			& 0.376		        & 0.634		      & 0.375	   	     & 0.628	              \\
		+ PCM					& 0.107		        & 0.171		      & 0.098		     &	0.145                  \\
		+ PCM + DIC				& \textbf{0.105}	& \textbf{0.126}  & \textbf{0.092}   &	\textbf{0.105}	        \\
		\bottomrule
	\end{tabular}
	\label{hard}
\end{table}

In this work, we propose a dynamic iterative correction (DIC) mechanism to provide a further correction on hard samples, which is verified effective in TABLE~\ref{components}. For a more intuitive observation, we conduct an analysis to reveal the effect of DIC on samples with different difficulty, as shown in TABLE~\ref{hard}. From the results, we have the following observations.
First, PCM can provide effective correction on all samples and improve the estimation performance.
Second, DIC works mainly on hard samples, \eg, samples with distance larger than 20m, and provide further improvement.

\subsubsection{Generalization to Different Lenses}
\begin{table}[!t]
	%\vspace{3mm}
	\caption{\textbf{Analysis on generalization to different lenses.} We trained our proposed modules on data collected via a 12mm len, while testing them on different data collected via a 12mm and a 16mm lens, respectively.}
	%\vspace{-6mm}
	\centering
	\renewcommand{\arraystretch}{1.3}
	\begin{tabular}{l|c|c|c|c}
		\toprule
		\multicolumn{1}{c|}{\multirow{2}{*}{Method}} & \multicolumn{2}{c|}{$12mm$}    & \multicolumn{2}{c}{$16mm$}				\\
		\cline{2-5}
				 		  & Abs Rel			& Sq Rel		& Abs Rel	    & Sq Rel		\\
		\hline\hline
		Baseline	 			& 0.494		      & 6.818		  & 0.430	   	  & 5.090	      \\
		+ PCM					& 0.121		      & 0.620		  & 0.199		  &	1.805          \\
		+ PCM + DIC				& \textbf{0.098}   & \textbf{0.401}  & \textbf{0.180} &	\textbf{3.186}	        \\
		\bottomrule
	\end{tabular}
	\label{lenses}
\end{table}

Since PCM and DIC need to be trained on data collected by predefined lenses, it is interesting to investigate the generalization ability of our method to testing data collected by different lenses. A promising generalization performance can make our method convenient to deploy on different UAVs efficiently after pretraining. Here, we collect a data subset via a $16mm$ len with similar distance distribution to our test subset, which is different from the $12mm$ len used in building UAVDE. As shown in TABLE~\ref{lenses}, our method can still bring significant improvement comparing to the baseline. However, a performance degradation occurs comparing to testing on in-distribution data (\ie, $12mm$), which remains as a direction for future research to improve the generalization ability.

\subsubsection{Generalization to Different Target UAVs}
\begin{table}[!t]
	%\vspace{3mm}
	\caption{\textbf{Analysis on generalization to different targets.} We train PCM and DIC on UAVDE dataset with a predefined target (M200 Orange) and evaluate them on different targets (M200 Orange, M200 Black and P600) based on different UAV detections.}
	%\vspace{-6mm}
	\centering
	\renewcommand{\arraystretch}{1.3}
	\begin{tabular}{c|l|c|c}
		\toprule
		UAV Detection                                 & \makecell[c]{Method}           & M200 Black		    & P600 		\\
	   \hline\hline
        \multirow{3}{*}{\makecell{Unfinetuned \\ Detector}}         & Baseline		     & 2.551				& 2.399	                 \\
		                                                          & + PCM			   & 6.466                & 2.815                   \\
  	                                                            & + PCM + DIC		 & \textbf{1.697}	    & \textbf{2.718}		  \\
	\hline
        \multirow{3}{*}{\makecell{GroundTruth \\ Detection}}      & Baseline		     & 0.434			    & 0.421	                     \\
		                                                         & + PCM			    & 0.140			       & 0.128                       \\
  	                                                         & + PCM + DIC		    & \textbf{0.092}	   & \textbf{0.124}		          \\
        \hline
        \multirow{3}{*}{\makecell{Finetuned \\ Detector}}         & Baseline		     & 0.421			    & 0.410	          \\
		                                                         & + PCM			    & 0.164                & 0.183           \\
  	                                                         & + PCM + DIC		    & \textbf{0.096}	   & \textbf{0.133}	\\
        \bottomrule
	\end{tabular}
	\label{targets_sum}
\end{table}

\begin{figure}[ht]
	\begin{center}
		\includegraphics[width=1.0\linewidth]{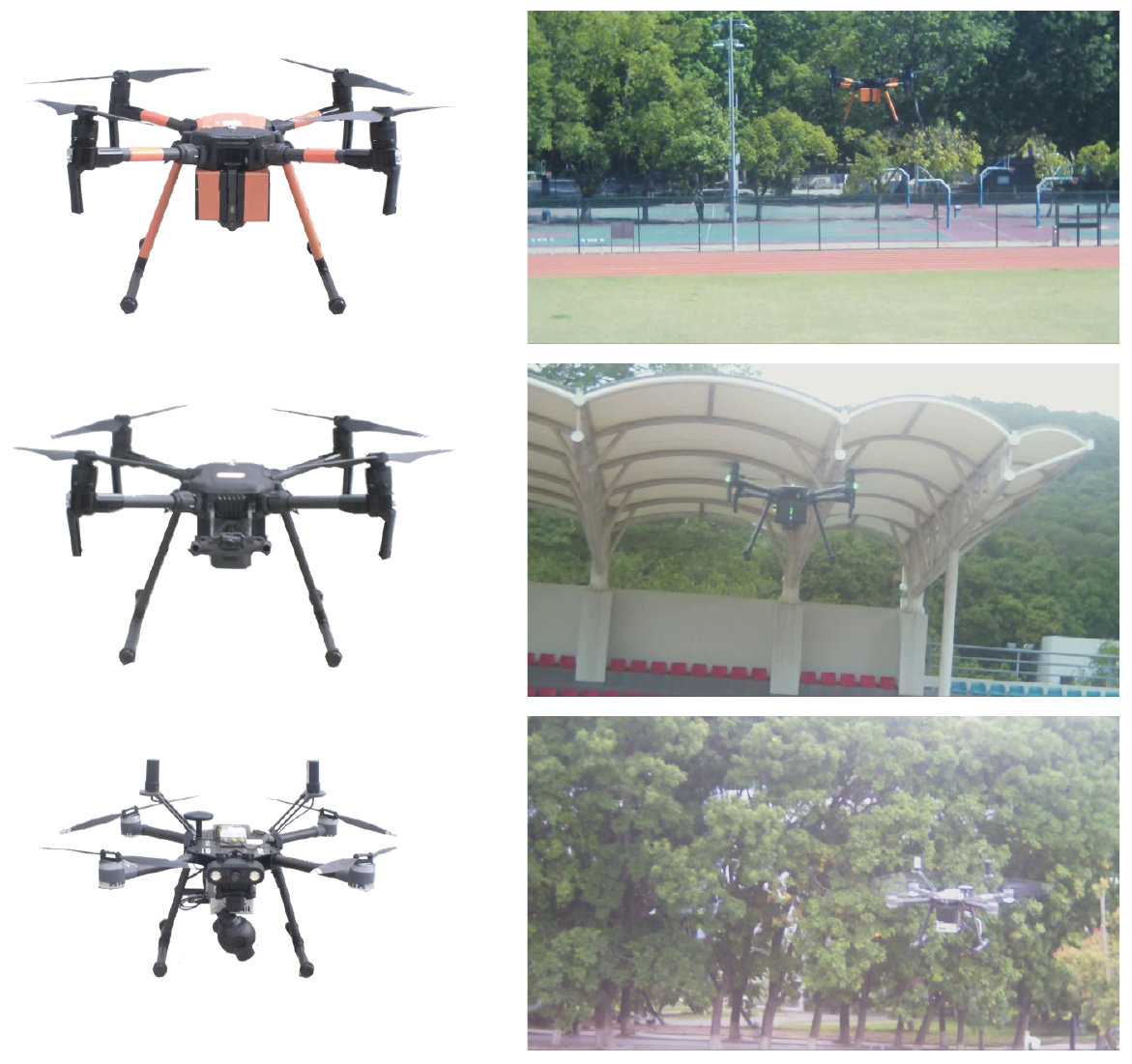}
	\end{center}
	%\vspace{-3mm}
	\caption{\textbf{The illustration of different target UAVs}. From top to bottom is M200 Orange, M200 black and P600, respectively. Best viewed in color.}
	%\vspace{-3mm}
	\label{uav_targets}
\end{figure}

Similarly, we also investigate whether our method can generalize to different target UAVs. Here, we further collect two data subsets of different target UAVs, including a UAV of same type but with different color (M200 Black) and a UAV of different type (P600), as shown in Fig.~\ref{uav_targets}. After pretraining on a predefined target UAV (M200 Orange) in the UAVDE dataset, we directly deploy our modules and the UAV detector on two out-of-distribution (OOD) subsets.
From the results in TABLE~\ref{targets_sum}, we can have the following observations. 
When directly deploying on unseen UAVs, the performance is unsatisfactory, due to a poor generalization ability to OOD samples.
However, when deploying our modules, \ie, PCM and DIC, on the groundtruth bounding box of unseen UAVs, a similar performance improvement can be obtained. The results demonstrate that the generalization issue is caused by the UAV detector instead of our proposed modules. 
To improve the generalization ability of detectors is an interesting and widely-studied topic~\cite{oza2023unsupervised,zhou2022domain,wang2022generalizing,yang2021generalized}, which is out of the scope of this paper.
To further confirm our inference, we collect a small amount of unseen UAVs samples to finetune the detector, \eg, less than 50 samples for each UAV type. Based on detections from the finetuned detector, our module can achieve similar performance. 

\subsubsection{Visualization on Correction Effect}
\begin{figure}[ht]
	\begin{center}
		\includegraphics[width=1.0\linewidth]{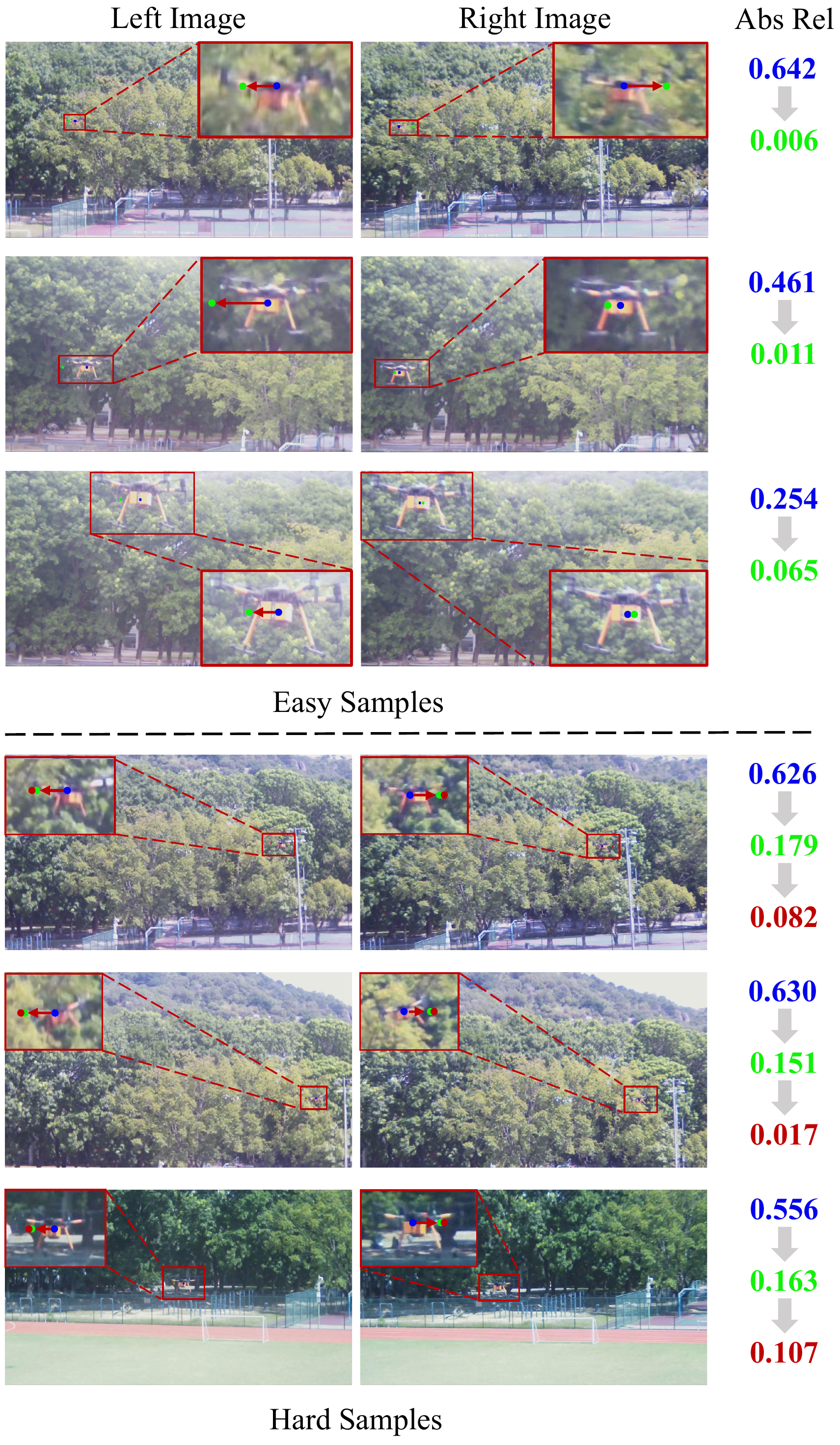}
	\end{center}
	%\vspace{-3mm}
	\caption{\textbf{Visualization of correction effect on easy and hard samples}. We visualize the observed and corrected positions of different samples for comparison. Here, we use dots with different colors for better visualization, \eg, blue for the observed position, green for the corrected position after PCM and red for the corrected position after DIC. The estimation performance of the sample is illustrated on the right. Best viewed in color and zoom in.}
	%\vspace{-3mm}
	\label{visualization}
\end{figure}

Here, we provide a visualization analysis on easy and hard samples from the test subset of UAVDE, which provides an intuitive observation on the effect of position correction, as shown in Fig.~\ref{visualization}. Specifically, we provide the trajectory of UAV position after correction and the improvement of estimation performance. It can be seen that our method can effectively tackle the position deviation issue.

%------------------------------------------------------------------------
\section{Conclusion} \label{sec:conclusion}
In this paper, we focus on the UAV distance estimation problem, which is practically important but rarely studied. To aid the study, we build a novel UAVDE dataset, and surprisingly find that the commonly used stereo triangulation paradigm does not work in UAV scenes. The main reason is the position deviation issue caused by the small baseline-to-depth ratio, large focal length and the vibrations on camera system, which are common in UAV scenes. To tackle this issue, we propose a novel position correction module (PCM) to explicitly predict the offset between the observed and actual positions of the target UAV, which is used for compensation in stereo triangulation calculation. Besides, we design a dynamic iterative correction (DIC) mechanism to further improve the correction effect on hard samples. Extensive experiments validate the effectiveness and superiority of our method.


%\section{References Section}
% You can use a bibliography generated by BibTeX as a .bbl file.
% The IEEEtran BibTeX style support page is:
 
\bibliographystyle{IEEEtran}
\bibliography{reference_new}
\end{document}